\documentclass{article}

\usepackage{algorithm}
\usepackage{algorithmicx}
\usepackage{algpseudocode}
\usepackage{mathtools}
\usepackage{url}
\usepackage{graphicx}

\usepackage[numbers]{natbib}

\usepackage{epstopdf}
\title{Improving energy efficiency and classification accuracy of neuromorphic chips by learning binary synaptic crossbars}

\author{Antonio Jimeno Yepes, Jianbin Tang\\
        antonio.jimeno@au1.ibm.com, jbtang@au1.ibm.com \\
        IBM Research Australia, Carlton, 3053, VIC, Australia}

\date{}

\begin{document}

\maketitle

\begin{abstract}

Deep Neural Networks (DNN) have achieved human level performance in many image analytics tasks but DNNs are mostly deployed to GPU platforms that consume a considerable amount of power.
Brain-inspired spiking neuromorphic chips consume low power and can be highly parallelized. However, for deploying DNNs to energy efficient neuromorphic chips the incompatibility between continuous neurons and synaptic weights of traditional DNNs, discrete spiking neurons and synapses of neuromorphic chips has to be overcome.
Previous work has achieved this by training a network to learn continuous probabilities and deployment to a neuromorphic architecture by random sampling these probabilities. An ensemble of sampled networks is needed to approximate the performance of the trained network.

In the work presented in this paper, we have extended previous research by directly learning binary synaptic crossbars.
Results on MNIST show that better performance can be achieved with a small network in one time step (92.7\% maximum observed accuracy vs 95.98\% accuracy in our work).
Top results on a larger network are similar to previously published results (99.42\% maximum observed accuracy vs 99.45\% accuracy in our work). More importantly, in our work a smaller ensemble is needed to achieve similar or better accuracy than previous work, which translates into significantly decreased energy consumption for both networks. 
Results of our work are stable since they do not require random sampling.
\end{abstract}

\section{Introduction}

Brain inspired neural networks can be deployed to low power and highly parallelized systems.
With recent advances in hardware platforms already available, there is an urge to provide software that fully exploits the potential of this new type of hardware.

Recent approaches in machine learning have explored training models constrained to binary weights~\cite{courbariaux2015binaryconnect} or low precision arithmetic~\cite{courbariaux2014low,cheng2015training,hwang2014fixed} and spiking neural networks~\cite{stromatias2015scalable}, achieving state of the art performance in image analytics tasks.
Recent work~\cite{esser2016convolutional} has used these ideas to implement constrained convolutional neural networks to learn discrete weights to enable high performance image analytics on TrueNorth~\cite{merolla2014million}.
These models have interesting properties since typically no multiplication modules are required. Lower power consumption is achieved by using only additions, which can benefit from implementing these models in configurable architectures such as FPGA or brain inspired computing systems.
These new models can enable low power brain-inspired computing analytics if efficiently translated to run on low power architectures.

In this work, we extend an existing DNN approach~\cite{esser2015backpropagation} to train a fully constrained synaptic crossbar.
Previous work trains a floating point crossbar connection model, which is sampled to obtain a discrete version that is deployed to a neurosynaptic chip.
We find that this sampling is problematic for some data sets such as electroencephalography (EEG) data, since the classification accuracy of the deployed network decreases.
To solve this problem, we propose a method that trains the previous network structure but learns a binary connection value instead of floating point values, which effectively means training trinary weights.
Results show that training binary crossbar connections instead of sampling their probability improves analytics performance for some MNIST networks and EEG data.

\subsection{Deployment hardware}

\begin{figure}[!ht]
  \centering
  \includegraphics[width=0.45\columnwidth]{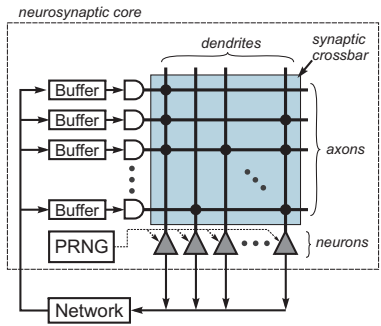}
  \caption{TrueNorth neurosynaptic-core~\cite{cassidy2013neuronmodel}}
  \label{fig:truenorth_core}
\end{figure}

IBM TrueNorth~\cite{merolla2014million} is a low power and highly parallelized brain inspired chip.
In its current implementation it is composed of 4,096 neurosynaptic cores.
Figure~\ref{fig:truenorth_core} shows the layout of one of these cores.
Each core has 256 input axons and 256 output neurons. 
Axons and neurons are connected by a configurable binary connection crossbar and neurons are highly configurable~\cite{cassidy2013neuronmodel}.

Programming for TrueNorth means writing core configurations and connections, which in a way could be seen as building a neural network.
A programming language has been developed to abstract the chip configuration~\cite{amir2013cognitive}.
In our case, the learned neural networks need to be translated into a valid TrueNorth configuration.
In previous work~\cite{esser2015backpropagation}, a constrained network is trained, which learns the crossbar connection probability.
To deploy the model into TrueNorth these probabilities are sampled into discrete synaptic crossbar connections.
We have extended this work to directly learn binary crossbar connections.

\section{Methods}

This section describes the classifier~\cite{esser2015backpropagation} on which this work is based on.
In this model, the aggregation function of each neuron input is defined in equation~\ref{eq:aggregation}, where $b_j$ is the bias term. In TrueNorth it can be set as the leak of the neuron and is used to guarantee that the neuron is stateless, so the neuron potential is reset at every tick.
$x_i$ is the input data on axon $i$, $c_{ij}$ is the crossbar connection status between axon $i$ and neuron $j$, and $s_{ij}$ is the synaptic strength between axon $i$ and neuron $j$.
During training, $x_i$ and $c_{ij}$ values will be within the range 0 to 1, while in the deployed network the values will be either 0 or 1.
$s_{ij}$ is defined as either [-1,1] or [-2,-1,1,2] as predefined by a template~\cite{esser2015backpropagation}.
The effective synaptic strength is defined by the element-wise multiplication of $c$ and $s$, with discrete weights defined as 1, -1 or 0 (no crossbar connection).

\begin{equation}
I_j = \sum_i x_i c_{ij} s_{ij} + b_j
\label{eq:aggregation}
\end{equation}

Equation~\ref{eq:aggregation} can be seen as a summation of weighted Bernoulli variables and the bias term.
Assuming independence between the inputs and with a large number of examples, it has been approximated using a Gaussian with mean and variance as shown in equation~\ref{eq:mean-variance}.

\begin{equation}
\begin{split}
\mu_j = b_j + \sum_i x_i c_{ij} s_{ij} \\
\sigma_j^2 = \sum_i x_i c_{ij} (1-x_i c_{ij}) s_{ij}^2
\end{split}
\label{eq:mean-variance}
\end{equation}

The parameters learned during training are the crossbar connections $c_{ij}$.
The other parameters are either fixed ($s_{ij}$) or are input data ($x_i$).

Given the neuron aggregation function $I_j$, the activation is defined in equation~\ref{eq:neuron-output}.
On the other hand, during training, the neuron output will be a value between 0 to 1 as defined by a Gaussian function~\cite{esser2015backpropagation}.

\begin{equation}
n_j =
  \begin{cases}
    1, & \mbox{if } I_j > 0 \\
    0, & \mbox{otherwise}
  \end{cases}
\label{eq:neuron-output}
\end{equation}

The described method allows defining fully connected layer networks, but TrueNorth cores have 256 inputs (axons) and 256 outputs (neurons), which constrains the size of fully connected layers.
When designing a network, the first layer, which processes the input data, is defined by blocks that tile to cover the input data. Block size and stride are defined by the number of pixels.
Since each one of these blocks maps to a TrueNorth core, the number of input axons required to map the input data per block is calculated as $({blockSize}^2 * numberChannels)$ and the maximum value is 256, which is the number of axons per core.
In the upper layers, block size and stride are defined by the number of cores (instead of pixels).
Output neurons of the previous layer are mapped to the input axons of the cores in the upper layer.

Training such a network from a TrueNorth perspective, effectively, consists of learning which connections are active in the crossbar and the bias term, which is used to set the leak value of the neurons.
The crossbar connections are learnt as connection probabilities in floating point, which needs to be translated to a discrete connection in the deployed crossbar.
In order to discretize the crossbar from the estimated probabilities, the probabilities are sampled to build an ensemble of classifiers.

\subsection{Binary crossbar learning}

The presented method has reported state of the art on MNIST~\cite{esser2015backpropagation}, in which the images are essentially black and white, with some degrees of gray.
Experiments using this method on EEG data have shown that the sampled deployment neural network has a lower performance than the floating point network~\cite{nurse2016acm}.

We propose to train binary crossbar connections directly versus the current method of learning connection probabilities that need to be sampled to develop a deployment network as shown in algorithm~\ref{alg:training}.

During training, in the forward propagation step $c^{bin}$ binary weights are estimated from the high precision $c$ matrix and are used to calculate the performance of the network.
During backpropagation, gradients are estimated using the $c^{bin}$ binary weights that are used to update the $c$ high precision weights.
Despite more refined methods, e.g. using a~\textit{hard tanh}, as proposed in~\cite{courbariaux2015binaryconnect}, we have used a simple rule to select the connection values from a high precision matrix:

\begin{equation}
c^{bin}_{ij} =
  \begin{cases}
    1, & \mbox{if } c_{ij} > 0.5 \\
    0, & \mbox{otherwise}
  \end{cases}
\end{equation}

\begin{algorithm}
\caption{Training algorithm iteration}
\label{alg:training}
\begin{algorithmic}
  \State{//Forward propagation}
	\For {$l=1:layers$}
	  \For {$c=1:layer(l).cores$}
	    \State Calculate core binary crossbar $c^{bin}_{ij} =
        \begin{cases}
          1, & \mbox{if } c_{ij} > 0.5 \\
          0, & \mbox{otherwise}
        \end{cases}$
	
	    \State Calculate core neuron outputs $I_j$ using $c^{bin}_{ij}$
		\EndFor
	\EndFor
	\State Estimate log loss in the last layer
	\State{//Backpropagation}
  \For {$l=1:layers$}
	  \For {$c=1:layer(l).cores$}
      \State Estimate core gradient values using $c^{bin}_{ij}$
	    \State Update core $c{ij}$ using estimated gradients
		\EndFor
	\EndFor
\end{algorithmic}
\end{algorithm}

\section{Results}

Experiments were run using a Caffe~\cite{jia2014caffe} implementation of the proposed method.
We have used two configurations for the axon type weights in the neuron configuration s=[-1,1] as used in previous work and as well the weights s=[-2,-1,1,2].

MNIST data set~\cite{lecun1998gradient} has been used in the experiments.
It contains handwritten digits from 0 to 9 with 60,000 examples for training and 10,000 samples for testing.
Images are 28x28 pixels, with one 8-bit gray scale channel.
Gray scale values have been normalized dividing them by 255.

Table~\ref{tab:networks} lists the networks used in the experiments.
MNIST network configurations are the same ones as described in previous work~\cite{esser2015backpropagation}.
There is a small network with 2 layers that require 5 TrueNorth cores and a larger network with 4 layers that require 30 TrueNorth cores.

\begin{table}[ht]
\begin{centering}
\begin{tabular}{ccc}
\hline
Name        & Layer & Definition                 \\
\hline
MNIST small & 1     & Block size 16 Stride 12    \\
            & 2     & Block size 2  Stride 1   \\
\hline						
MNIST large & 1     & Block size 16 Stride 4 \\
            & 2     & Block size 2  Stride 1 \\
						& 3     & Block size 2  Stride 1 \\
						& 4     & Block size 2  Stride 1 \\
\hline
\end{tabular}
\par
\end{centering}
\caption{Networks used in the experiments.
         MNIST small has 2 layers and requires 5 TrueNorth cores.
         MNIST large has 4 layers and requires 30 TrueNorth cores.}
\label{tab:networks}
\end{table}

Performance of the deployed trained models has been measured using classification accuracy, which is calculated by dividing the number of correctly predicted instances by all instances.
Data is converted into spikes before sending it to the TrueNorth chip using a rate code scheme.
Since rate code is used and there is no need to sample from the trained network crossbar probabilities, results are stable (e.g. repetitions of the experiments provide the same result) compared to previous methods.

Training has been carried out using batches of 100 images for each iteration.
The initial learning rate is 0.1, which is multiplied by 0.1 after each 1 million iterations.
2 million iterations were used for training.

Experiments have been carried out with and without data augmentation.
To perform data augmentation, random changes to the images have been done in each epoch.
We have used two augmentation sets of values.
For the first one (Aug1), we used a maximum rotation of 7.5 degrees, a maximum shift of 2.5 and a maximum rescale of 7.5\%.
For the second one (Aug2), we used a maximum rotation of 15 degrees, a maximum shift of 5 and a maximum rescale of 15\%, which is the same configuration used in previous work.

Table~\ref{tab:results-mnist-small} shows the result with the small MNIST networks (5 cores) with and without data augmentation and with an increasing number of ticks.
The performance is higher compared to previously published work with 1 tick, i.e. 0.927~\cite{esser2015backpropagation} vs 0.9598 (cf. table~\ref{tab:results-mnist-small}).

\begin{table}[ht]
\begin{centering}
\begin{tabular}{c|rrr|rrr}
\hline
     &\multicolumn{3}{c|}{s=[-2,-1,1,2]}&\multicolumn{3}{c}{s=[-1,1]}\\
\hline
Ticks&No aug&Aug1	&Aug2 &No aug&Aug1 &Aug2  \\
\hline
1    &95.84 &94.92&91.18&95.98 &95.75&92.11 \\
2    &97.12 &96.69&93.79&96.96 &97.10&94.40 \\
4    &97.51 &97.49&95.00&97.55 &97.60&95.75 \\
8    &97.61 &97.72&95.45&97.50 &97.93&96.21 \\
16   &97.70 &98.03&95.86&97.71 &98.07&96.36 \\
32   &97.71 &98.00&95.75&97.73 &98.09&96.30 \\
64   &97.74 &98.09&95.96&97.80 &98.17&96.55 \\
\hline
\end{tabular}
\par
\end{centering}
\caption{MNIST small network results using data augmentation (Aug1, Aug2) and not using it (No aug).
}
\label{tab:results-mnist-small}
\end{table}

Table~\ref{tab:results-mnist-large} shows the performance on the large network (30 cores). 
With a low number of ticks, the performance is better compared to previously reported results.

\begin{table}[ht]
\begin{centering}
\begin{tabular}{c|rrr|rrr}
\hline
     &\multicolumn{3}{c|}{s=[-2,-1,1,2]}&\multicolumn{3}{c}{s=[-1,1]}\\
\hline
Ticks&No aug&Aug1	&Aug2 &No aug&Aug1 &Aug2  \\
\hline
1	   &93.91 &96.88&97.27&93.54&96.82 &96.77 \\
2	   &96.33 &98.10&98.18&95.62&98.32 &98.38 \\
4	   &97.28 &98.79&98.81&97.06&98.89 &99.06 \\
8	   &97.85 &98.90&98.75&97.66&99.16 &99.20 \\
16	 &98.01 &99.09&98.96&97.79&99.29 &99.24 \\
32	 &97.96 &99.20&99.07&98.02&99.23 &99.26 \\
64	 &98.13 &99.19&99.05&97.94&99.23 &99.29 \\
\hline
\end{tabular}
\par
\end{centering}
\caption{MNIST large network results using data augmentation (Aug1, Aug2) and not using it (No aug).
}
\label{tab:results-mnist-large}
\end{table}

Results for an ensemble of trained classifiers have been reported in table~\ref{tab:results-mnist-large-augmentation}.
Instead of sampling the network as in previous work, the networks have been trained with different augmentation values and neuron axon type weights.
Ensembles show an improvement of the results reported in table~\ref{tab:results-mnist-large} and provide a small improvement over previous work, in which a maximum of 99.42\% was observed using stochastic code and sampling the crossbar from a trained connection probability.
In this work we obtain 99.45\% using rate code and an ensemble of a fixed set of networks, which is more stable.
Our reported results rely on just 4 trained networks and multiple ticks, which indicates that additional trained networks could improve these results.

\begin{table}[ht]
\begin{centering}
\begin{tabular}{crr}
\hline
Ticks&Small Network&Large Network\\
\hline
1		 &	98.12 & 99.11\\
2		 &	98.40 & 99.27\\
4		 &	98.50 & 99.33\\
8		 &	98.63 & 99.38\\
16	 &	98.53 & 99.40\\
32	 &	98.60 & 99.45 \\
64	 &	98.57 & 99.41 \\
\hline
\end{tabular}
\par
\end{centering}
\caption{Results on MNIST for an ensemble of trained classifiers based on different augmentation values and neuron axon type weights.}
\label{tab:results-mnist-large-augmentation}
\end{table}

\section{Discussion}

Results on MNIST show that the performance of the proposed method is similar or better than previously reported work depending on the configuration.
In the case of the smallest network for MNIST, the proposed method provides significantly better results.
Results on EEG data show that the performance of the training network increased from 76\%~\cite{nurse2016acm} to 84\% (same performance in deployment network), which is close to state-of-the-art performance obtained using unconstrained fully connected layers~\cite{nurse2015generalizable}.

The use of several ticks for encoding input data shows that a plateau is reached after 8 ticks for MNIST, and using additional ticks to encode data does not significantly increase performance.
The performance of the large network ensemble is similar to the performance of the floating point implementation presented in previous work.

\begin{figure}[htb]
\centering
  \begin{tabular}{@{}ccc@{}}
		\multicolumn{3}{c}{Neuron weight configuration s=[-1,1]} \\
    \includegraphics[width=.30\textwidth]{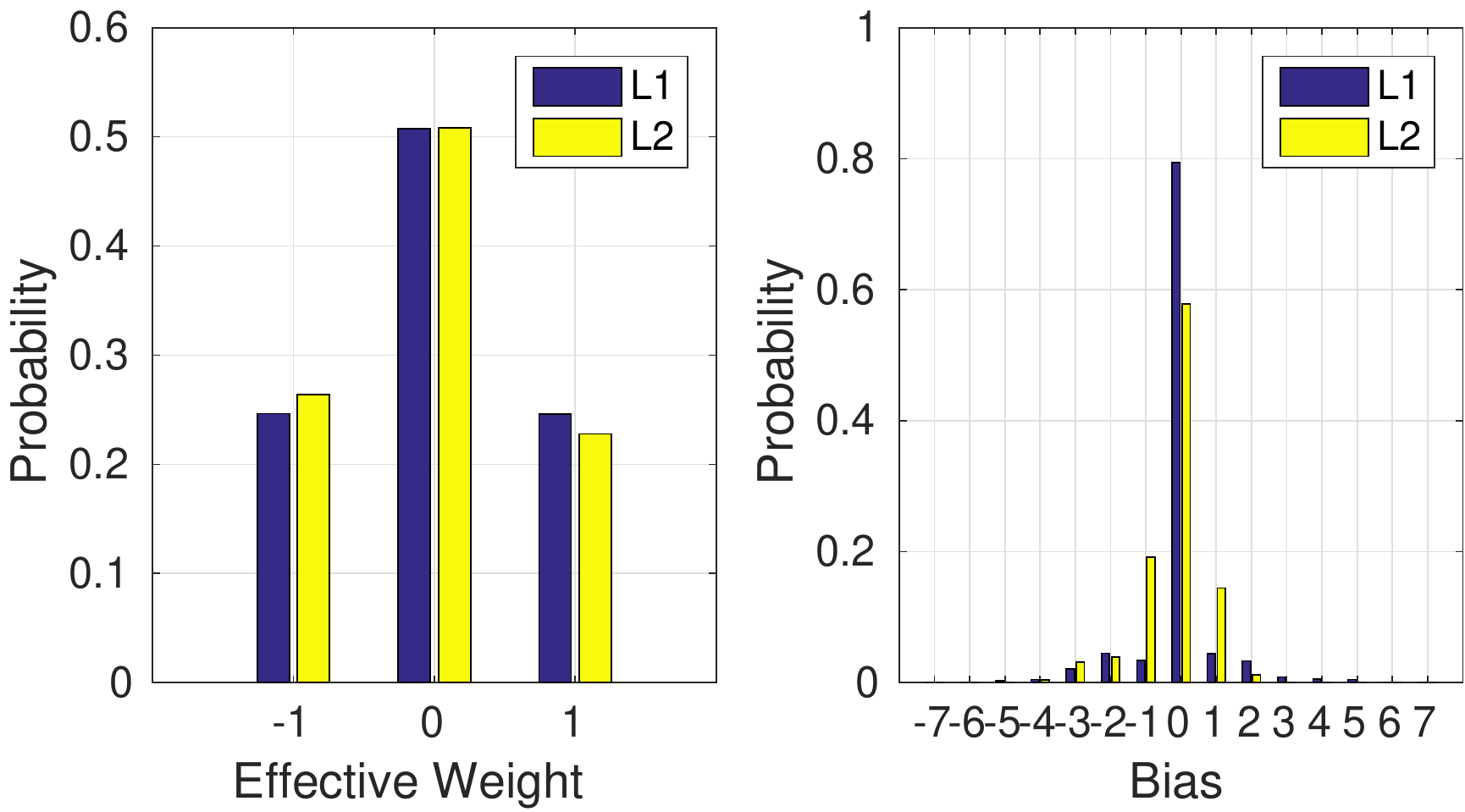} &
    \includegraphics[width=.30\textwidth]{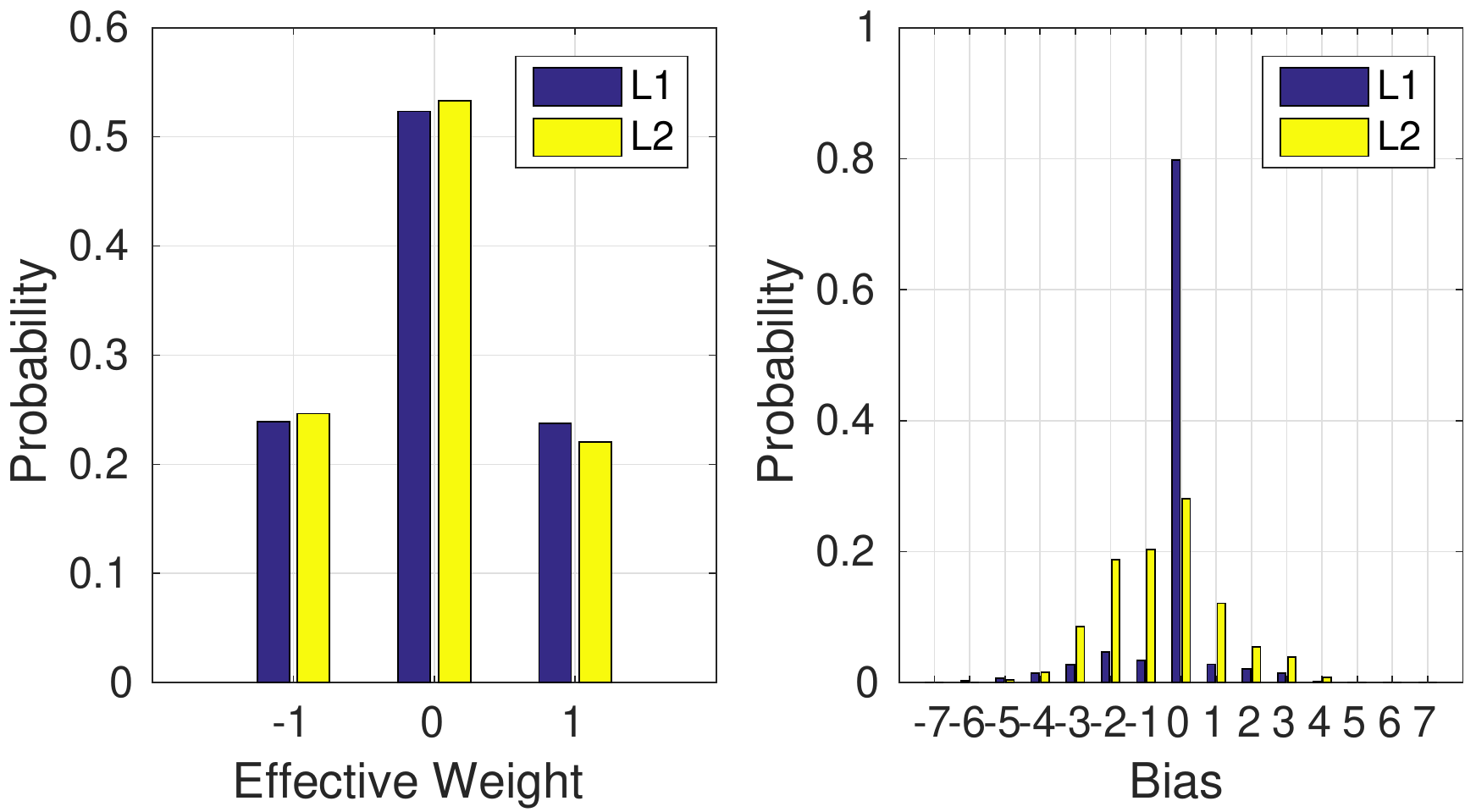} &
    \includegraphics[width=.30\textwidth]{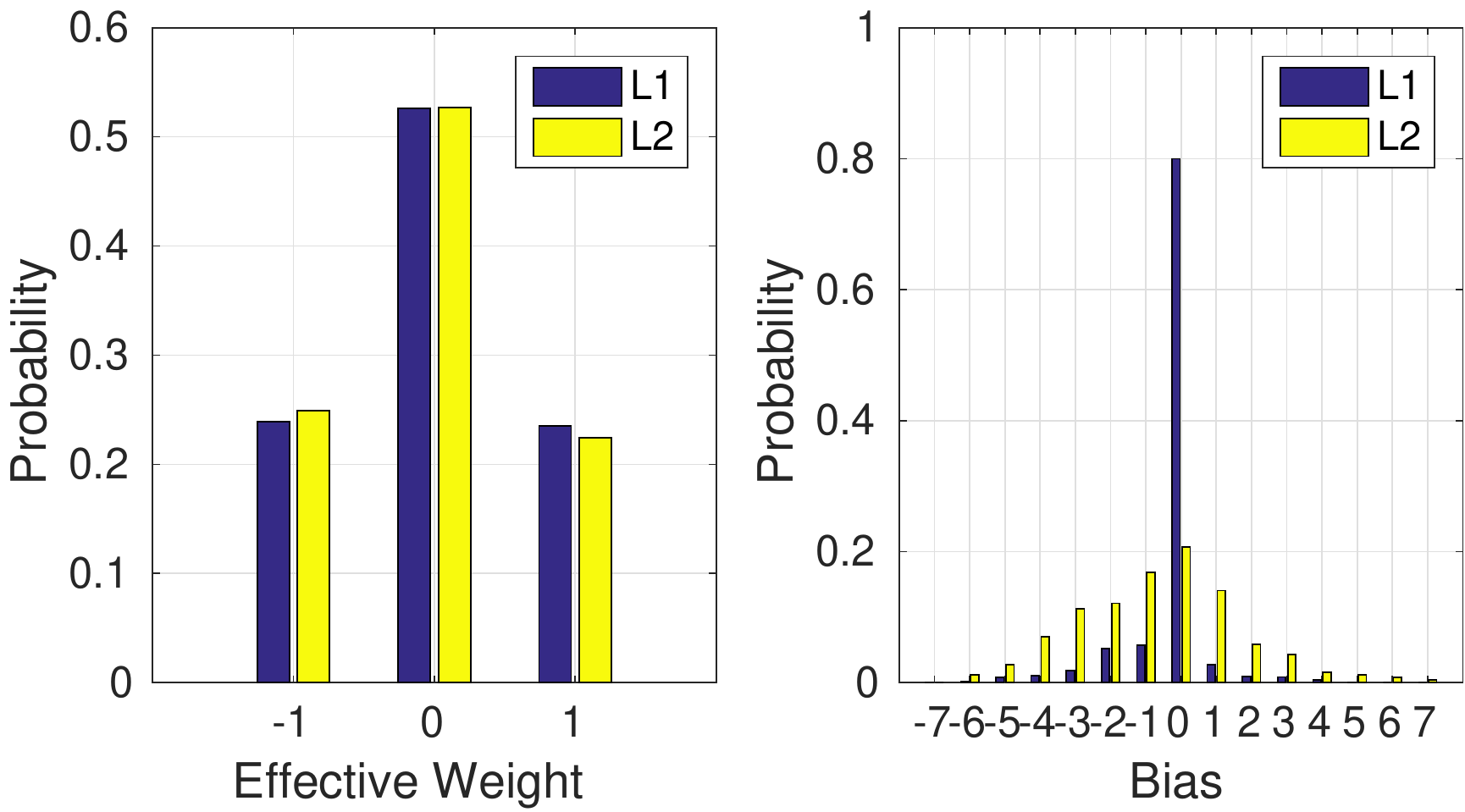} \\
		No augmentation&
		Aug1&
		Aug2\\
    \multicolumn{3}{c}{Neuron weight configuration s=[-2,-1,1,2]} \\
    \includegraphics[width=.30\textwidth]{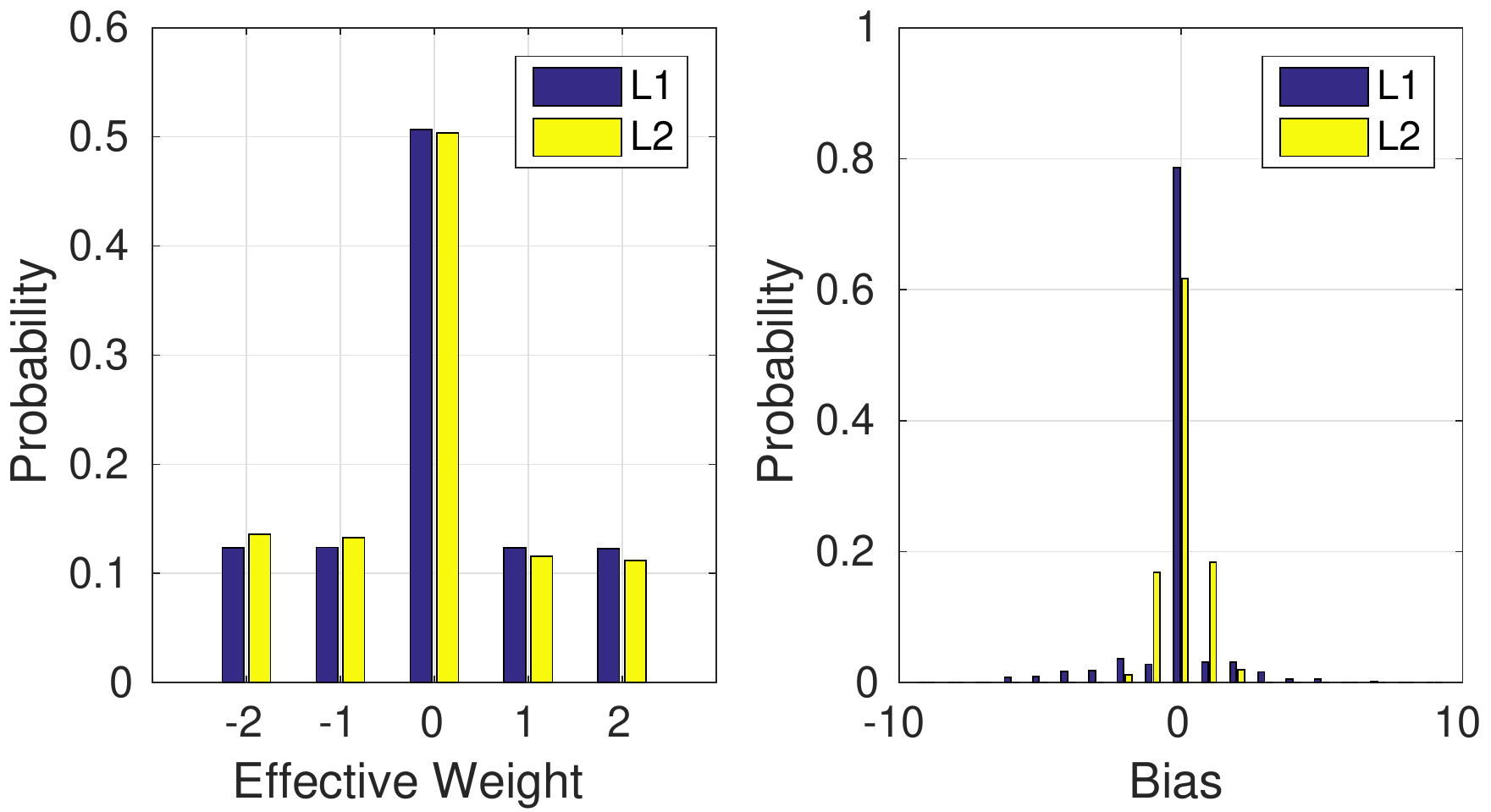} &
    \includegraphics[width=.30\textwidth]{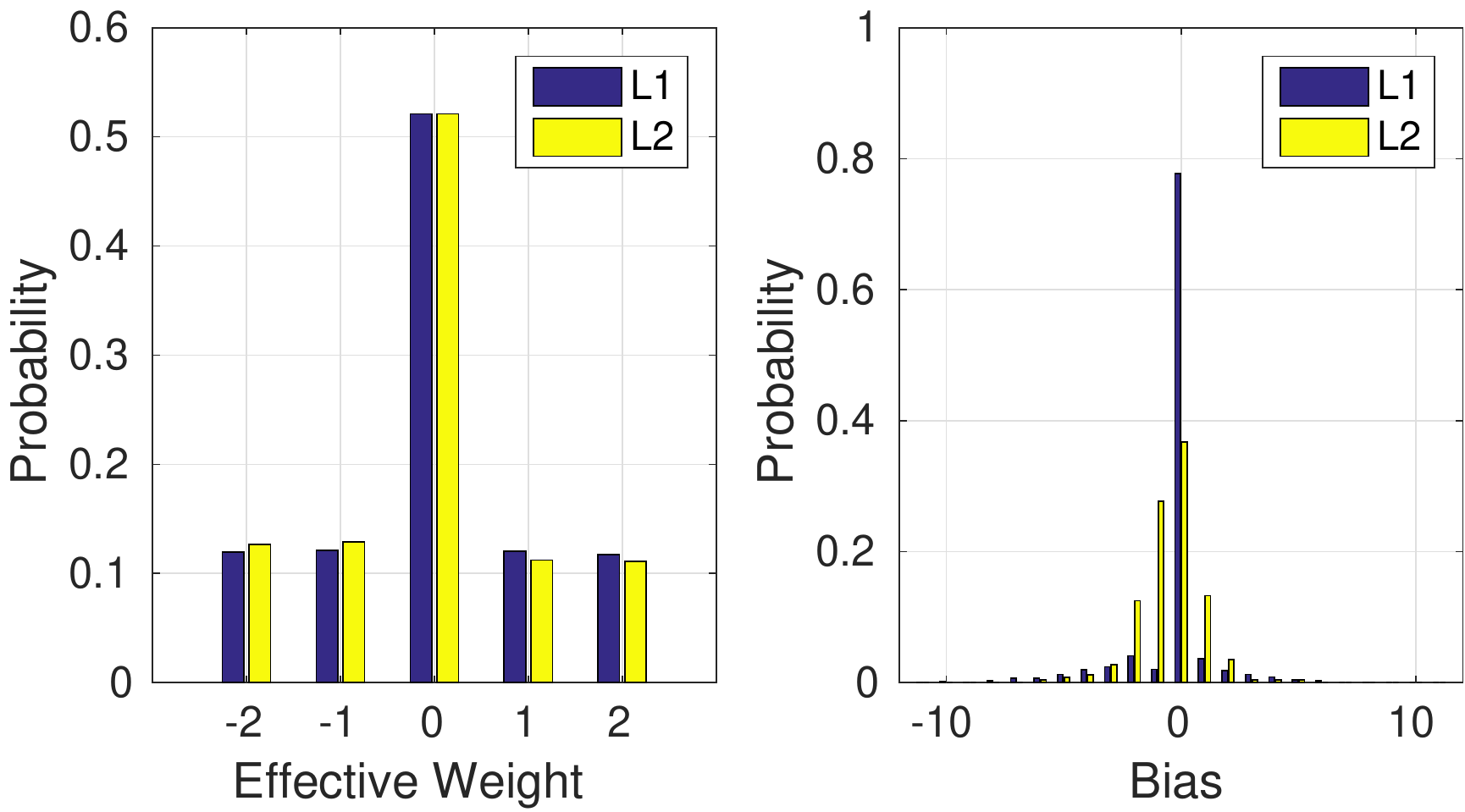} &
    \includegraphics[width=.30\textwidth]{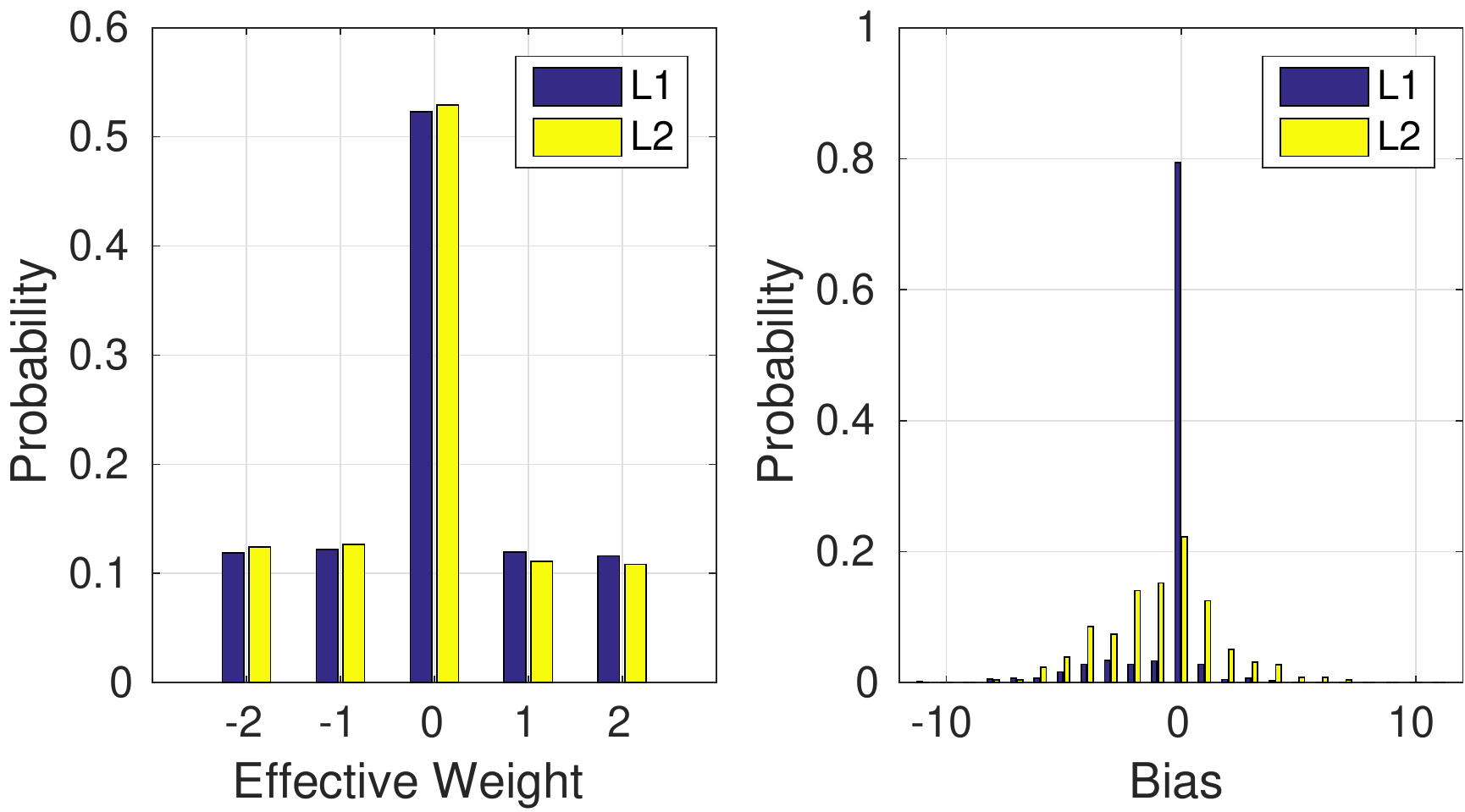} \\
		No augmentation&
		Aug1&
		Aug2\\
  \end{tabular}
  \caption{Effective weights and bias values for the large network (2 layers, 5 cores).
	         Two weight configurations have been used s=[-1,1] and s=[-2,-1,1,2].
					 The networks have been trained without augmentation (No augmentation) and with different augmentation values (Aug1, Aug2).
					}
	\label{fig:small-network-details}
\end{figure}

\begin{figure}[htb]
\centering
  \begin{tabular}{@{}ccc@{}}
		\multicolumn{3}{c}{Neuron weight configuration s=[-1,1]} \\
    \includegraphics[width=.33\textwidth]{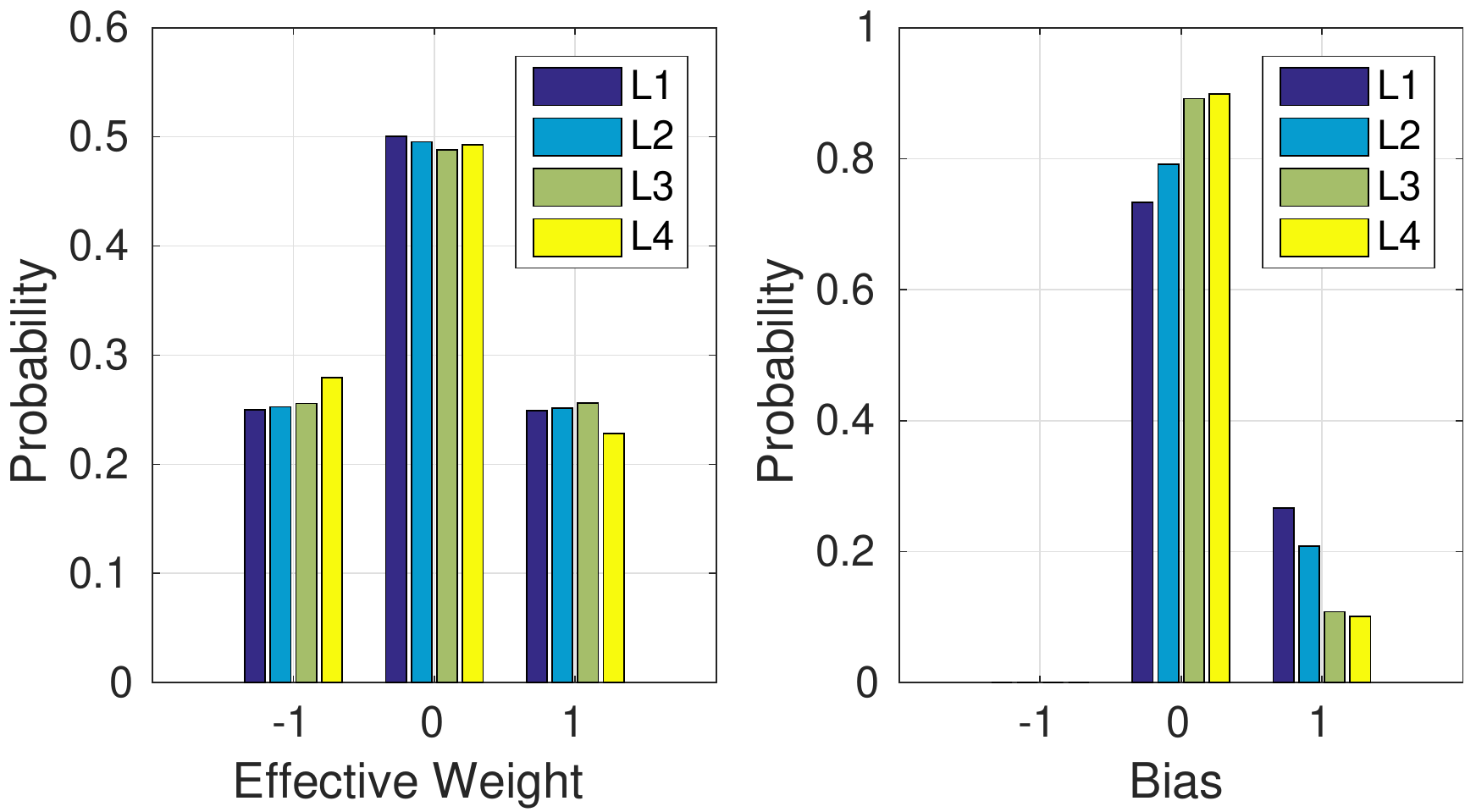} &
    \includegraphics[width=.33\textwidth]{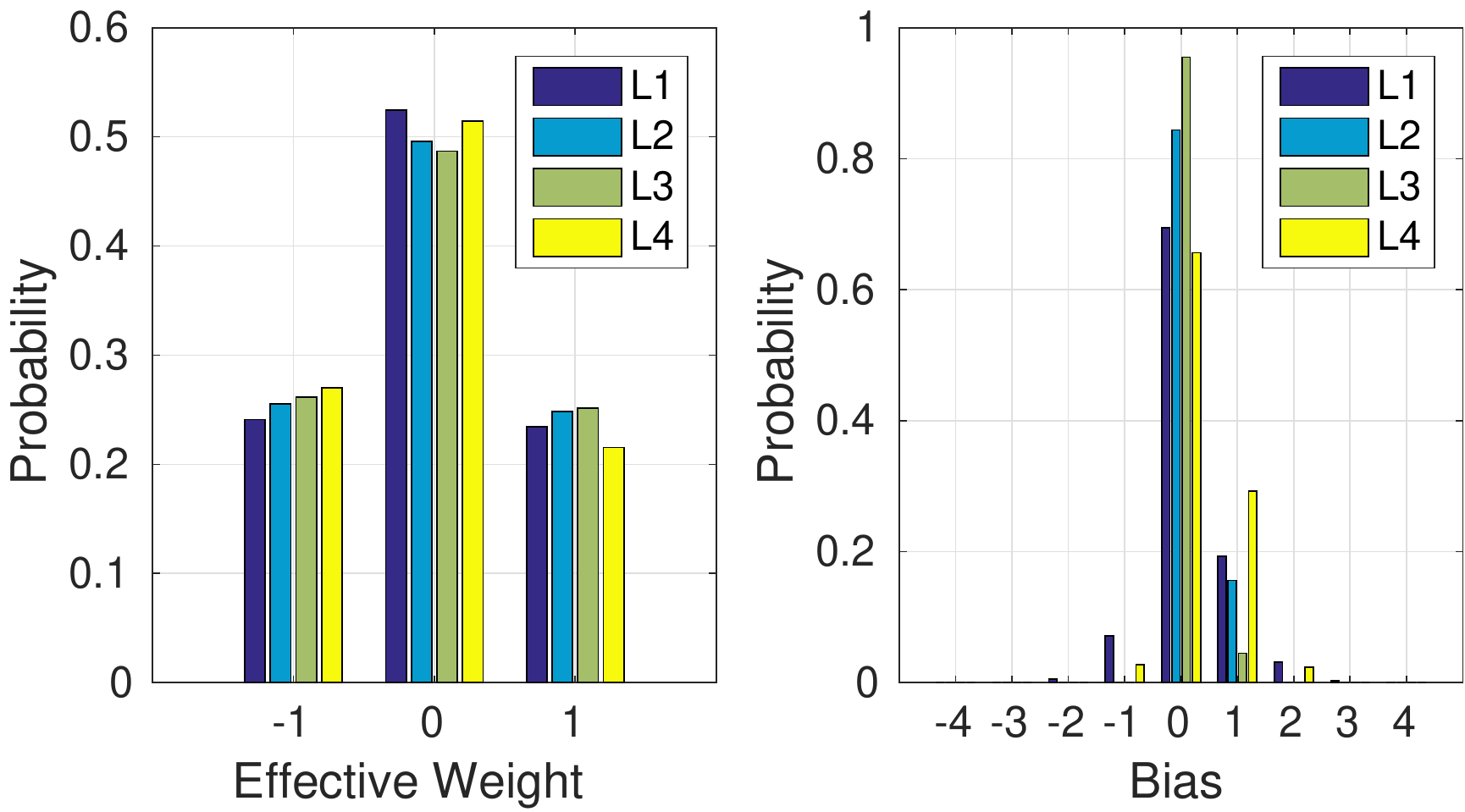} &
    \includegraphics[width=.33\textwidth]{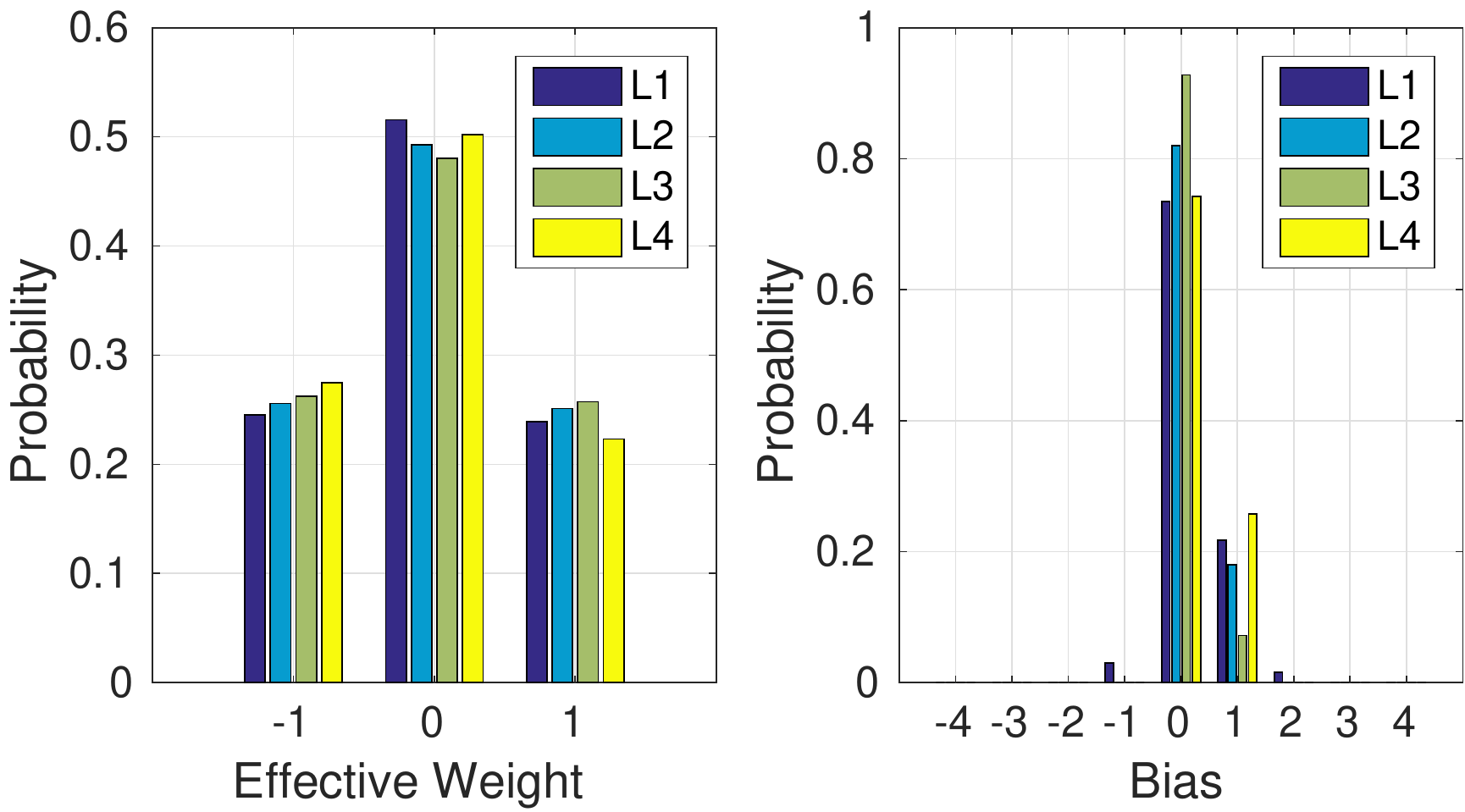} \\
		No augmentation&
		Aug1&
		Aug2\\
    \multicolumn{3}{c}{Neuron weight configuration s=[-2,-1,1,2]} \\
    \includegraphics[width=.33\textwidth]{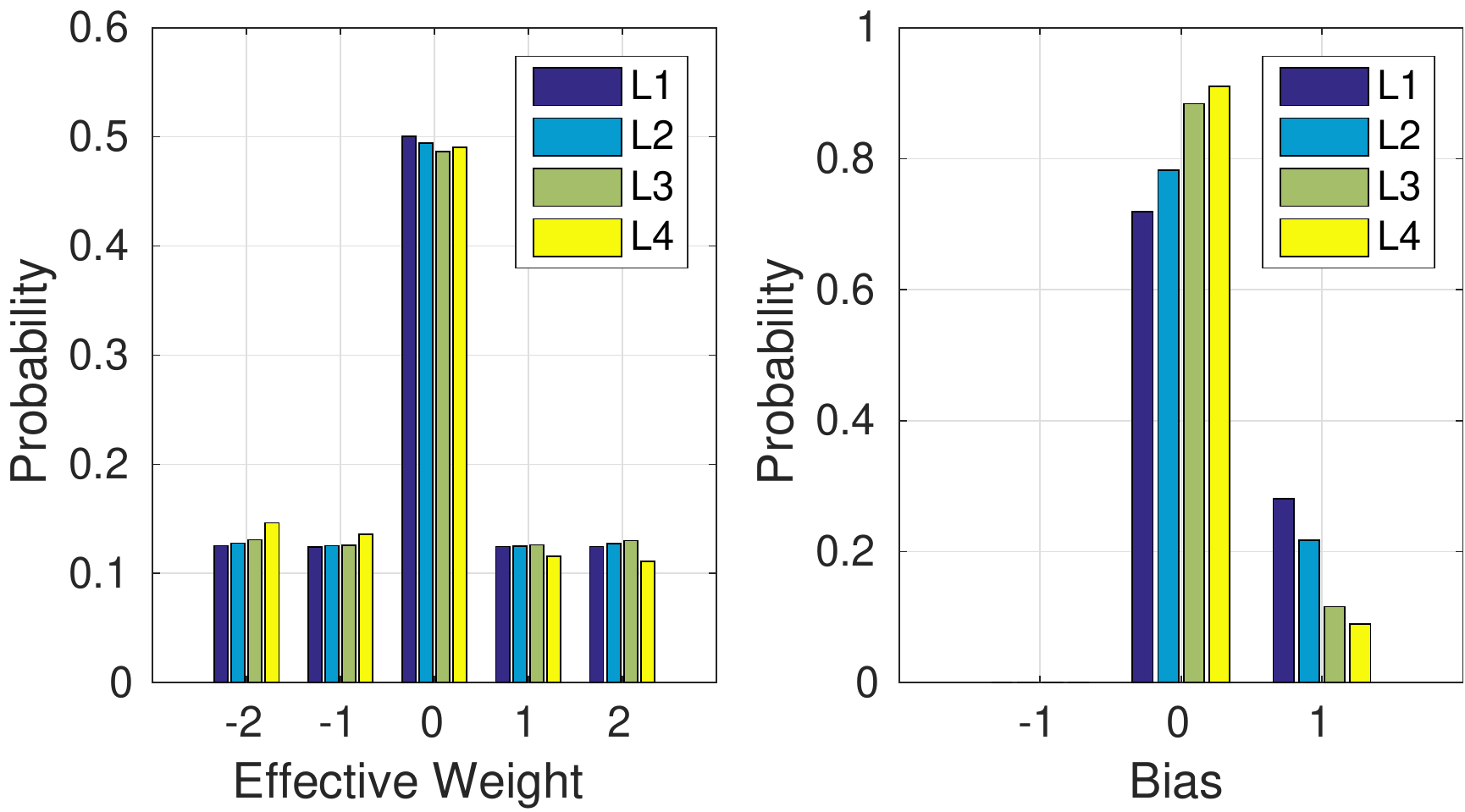} &
    \includegraphics[width=.33\textwidth]{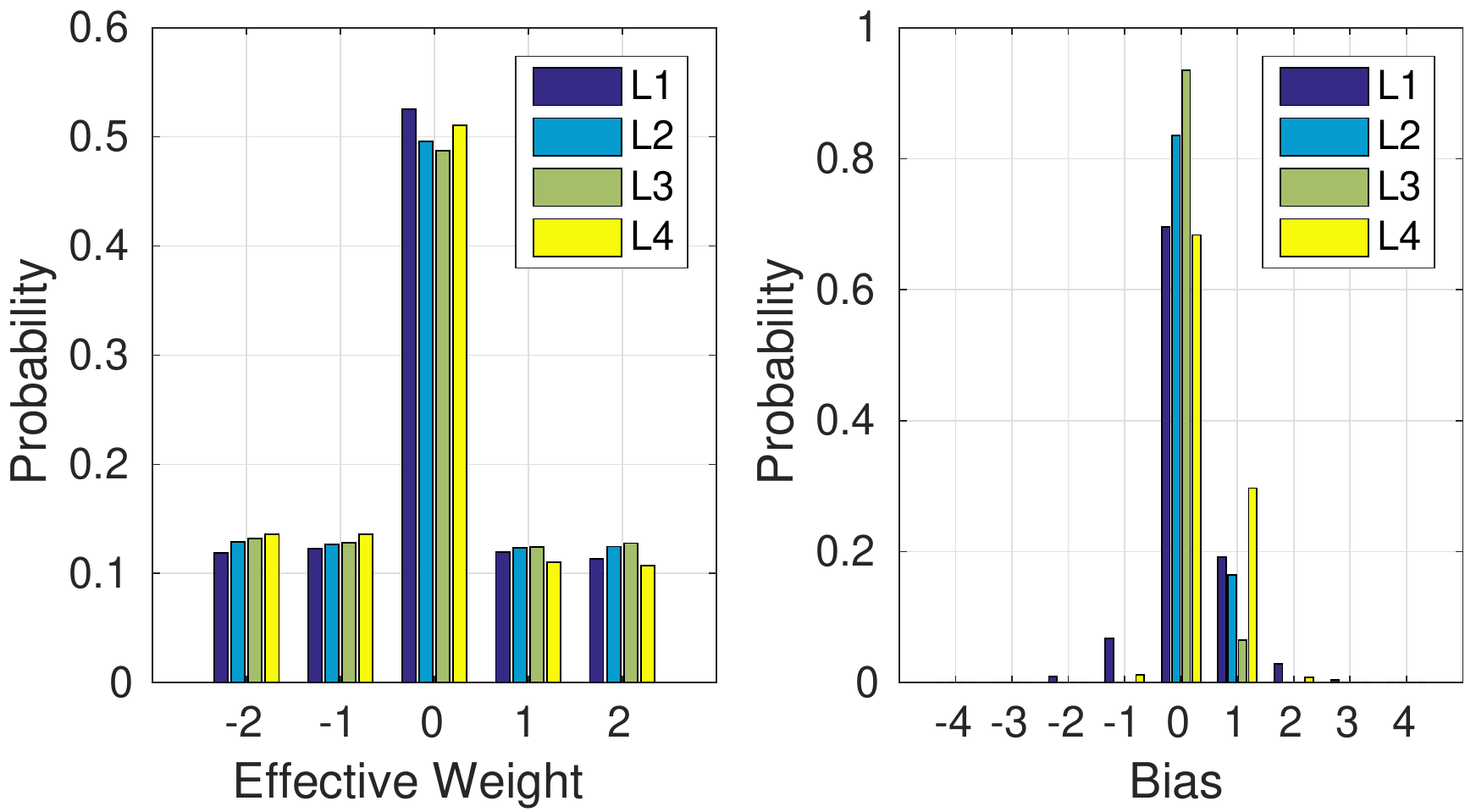} &
    \includegraphics[width=.33\textwidth]{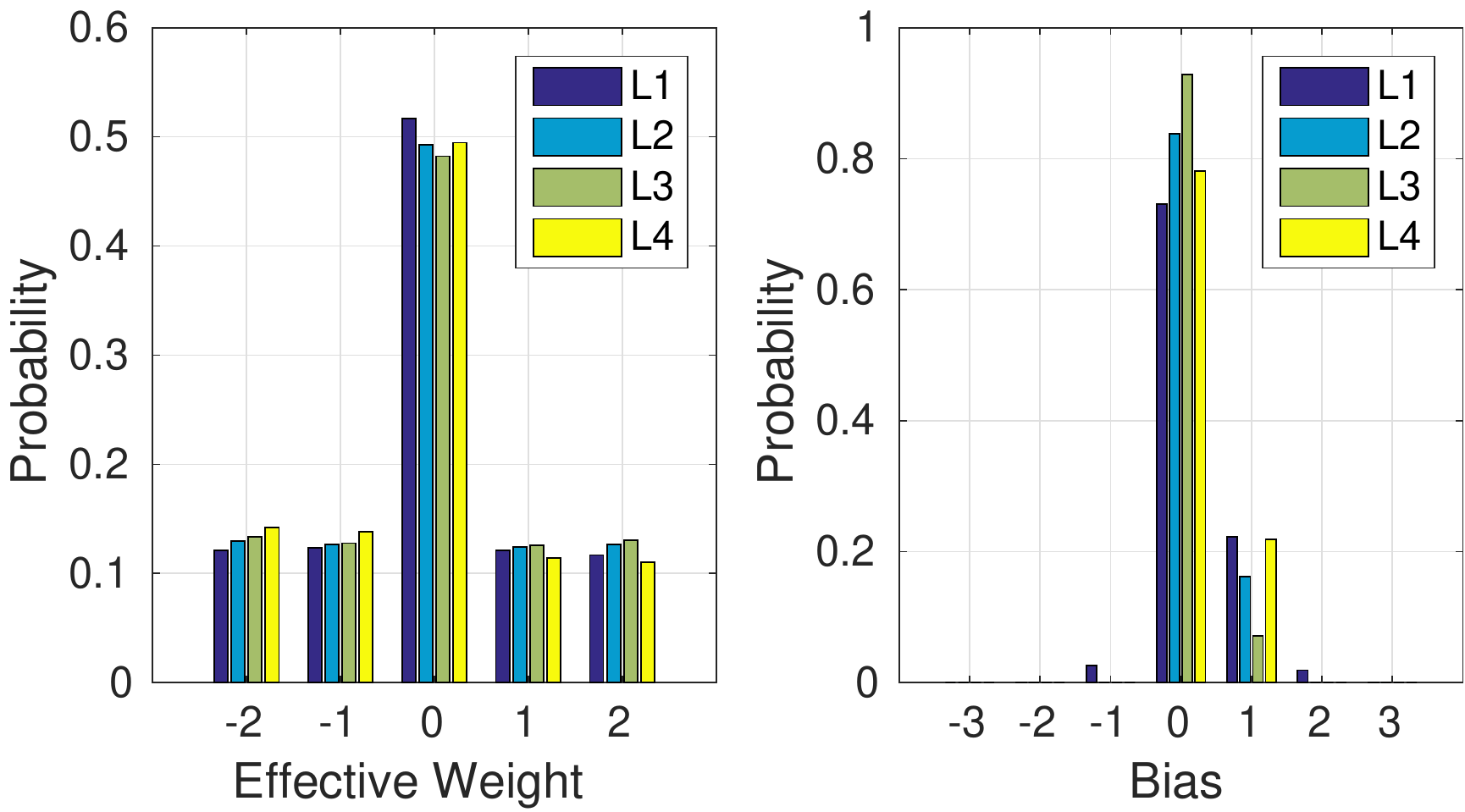} \\
		No augmentation&
		Aug1&
		Aug2\\
  \end{tabular}
  \caption{Effective weights and bias values for the large network (4 layers, 30 cores).
	         Two weight configurations have been used s=[-1,1] and s=[-2,-1,1,2].
					 The networks have been trained without augmentation (No augmentation) and with different augmentation values (Aug1, Aug2).
					}
	\label{fig:large-network-details}
\end{figure}

Figures~\ref{fig:small-network-details} and~\ref{fig:large-network-details} show the distribution of effective weights and biases for the small and large network respectively.
Effective weights are obtained by multiplying the binarized crossbar connections $c^{bin}_{ij}$ and the neuron axon type weights $s_{ij}$.
The graph shows the values for each one of the layers.
Half of the effective weights have a value of zero, which means that half of the possible crossbar connections are disconnected.
Independent of the weight configuration and network, the other weights are equally distributed.
As shown in figures \ref{fig:small-network-details} and~\ref{fig:large-network-details}, most of the bias values are zero.
In the small network, biases are distributed around zero, while in the large network, weights are mostly zero while some bias values in a smaller proportion are one.

We have evaluated two neuron axon type weights (s=[-1,1] and s=[-2,-1,1,2]).
Results show that results are not significantly better for the s=[-1,1] configuration and this result indicates that effective binary weights (-1,0,1) are sufficient for this task.

\begin{figure}[htb]
  \centering
  \includegraphics[width=0.8\textwidth]{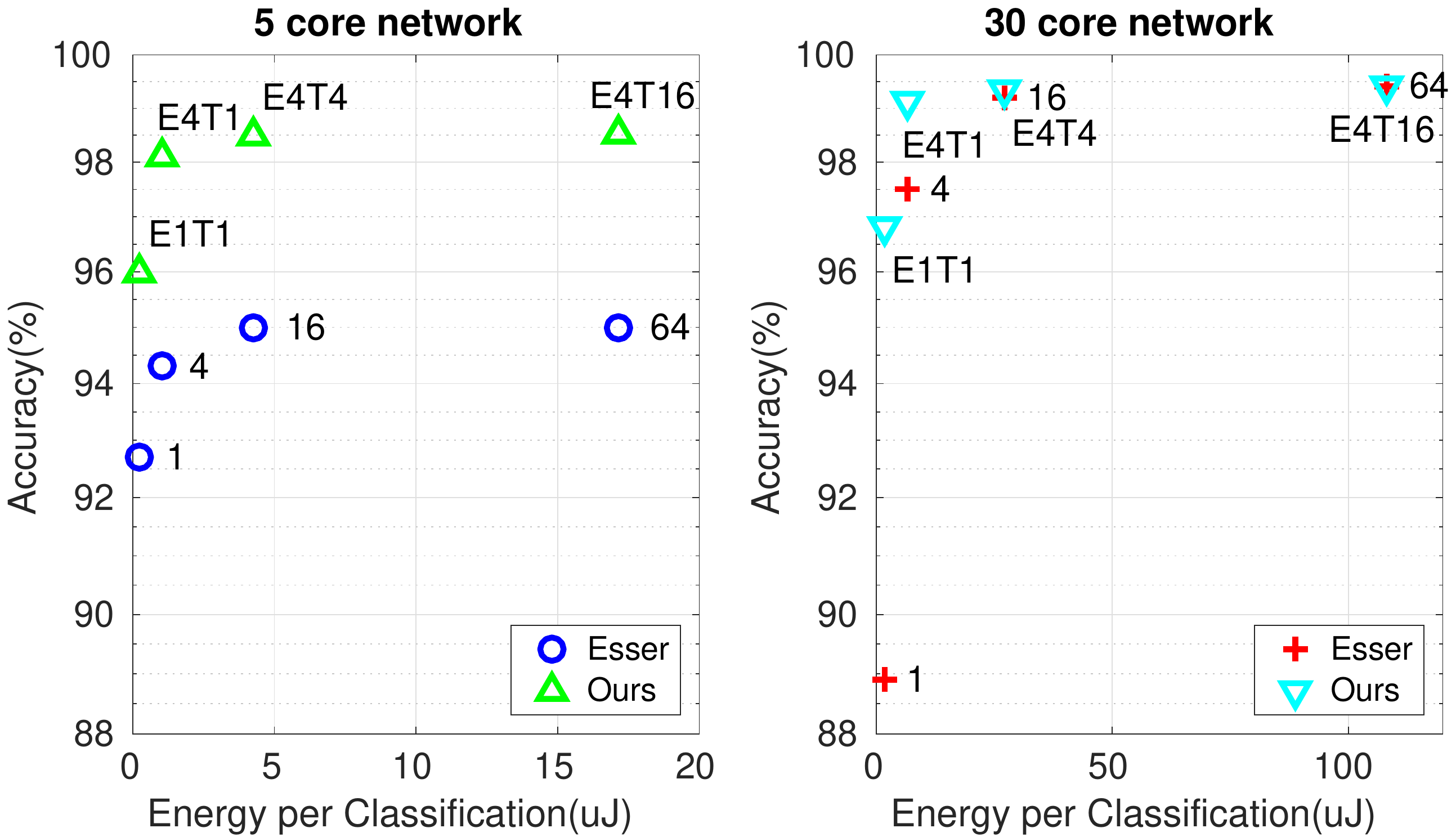}
  \caption{Energy (uJ) versus accuracy for the 5 core and 30 core networks.
	         Results are compared against previous work~\cite{esser2015backpropagation} (Esser) for similar energy configuration.
					 Previous work results are plotted according to ensemble size.
					 For our work, we show results by ensemble size (E) and number of ticks used to encode the data using rate code (T).
					}
	\label{fig:energy-vs-accuracy}
\end{figure}

Figure~\ref{fig:energy-vs-accuracy} shows the energy consumption on TrueNorth for several network configurations, which is compared with the energy consumption reported in previous work~\cite{esser2015backpropagation} (Esser).
In most configurations with similar energy consumption, the extension proposed in our work provides higher accuracy.
Ours and previous work seem to reach a plateau around 99.4\% with similar energy requirements.

\section{Conclusions and future work}

We have shown that it is possible to train constrained fully connected layers learning binary crossbar connections, which provides better or similar performance compared to previous work on the MNIST data set.
These results have interesting applications since no sampling of the crossbar connections needs to be done to deploy the system.
Results on EEG data sets which were previously analysed through the original method yield an 84\% classification accuracy using the new approach.

The outcome of this research provides interesting insights into the possibility of learning network parameters that easily translate into deployment networks for brain-inspired chips.
We will explore feasibility of expanding the introduced new methodology to train other kinds of existing deep learning algorithms including convolutional neural networks and recurrent neural networks towards running existing state of the art algorithms on brain-inspired architectures in real time and with extreme low power consumption.

\section{Acknowledgment}
The authors would like to thank the SyNAPSE team at the IBM Almaden Research Center for their support in providing the Caffe code.
The authors thank Dr. Isabell Kiral-Kornek, Dr. Stefan Harrer and Dr. Benjamin S. Mashford (IBM Research - Australia) for support and fruitful discussions.

\bibliographystyle{plain}
\bibliography{bibliography}

\end{document}